\documentclass{article}
\usepackage{ijcai18}

\usepackage{times}
\usepackage{xcolor}
\usepackage{soul}
\usepackage[utf8]{inputenc}
\usepackage[small]{caption}
\usepackage{graphicx} 
\usepackage{amsfonts}
\usepackage{amsmath}
\usepackage{url}
\usepackage{multirow}

\title{Facial Aging and Rejuvenation by Conditional Multi-Adversarial Autoencoder with Ordinal Regression}

\author{
Haiping Zhu$^1$,
Qi Zhou$^1$,
Junping Zhang$^1$,
James Z. Wang$^2$
\\ 
$^1$ Shanghai Key Lab of Intelligent Information Processing, \\
School of Computer Science, Fudan University, Shanghai 200433, China \\
$^2$ College of Information Sciences and Technology, \\
The Pennsylvania State University, University Park, PA 16802, USA \\
\{hpzhu14, qizhou15, jpzhang\}@fudan.edu.cn,
jwang@ist.psu.edu
}

\begin{document}

\maketitle
\begin{abstract}
Facial aging and facial rejuvenation analyze a given face photograph to predict a future look or estimate a past look of the person. To achieve this, it is critical to preserve human identity and the corresponding aging progression and regression with high accuracy. However, existing methods cannot simultaneously handle these two objectives well. We propose a novel generative adversarial network based approach, named the Conditional Multi-Adversarial AutoEncoder with Ordinal Regression (CMAAE-OR). It utilizes an age estimation technique to control the aging accuracy and takes a high-level feature representation to preserve personalized identity. Specifically, the face is first mapped to a latent vector through a convolutional encoder. The latent vector is then projected onto the face manifold conditional on the age through a deconvolutional generator. The latent vector preserves personalized face features and the age controls facial aging and rejuvenation. A discriminator and an ordinal regression are imposed on the encoder and the generator in tandem, making the generated face images to be more photorealistic while simultaneously exhibiting desirable aging effects. Besides, a high-level feature representation is utilized to preserve personalized identity of the generated face.  Experiments on two benchmark datasets demonstrate appealing performance of the proposed method over the state-of-the-art.
\end{abstract}

\section{Introduction}\label{intro}
Facial aging and facial rejuvenation, also referred to as age progression and age regression, aim to render a face photograph with natural aging and rejuvenating effects on the individual's face~\cite{fu2010age}. It has broad applications, including facial appearance prediction, cross-age face recognition~\cite{park2010age}, finding missing children, and movie entertainment~\cite{wang2016recurrent}.
\begin{figure}[!ht]
\centering
\includegraphics[width=1.0\linewidth,height=0.65\linewidth, clip=true, trim=0 0 0 0]{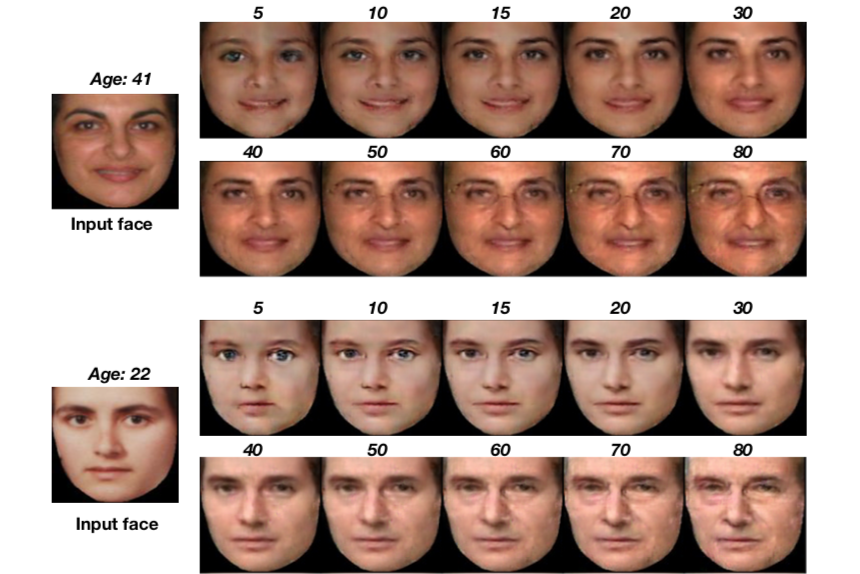}
\caption{Example simulation results of our aging and rejuvenation process. The first image of each row is the provided face photograph, with the actual age marked. The other images are machine-generated face images for the specified ages of the same person.
}
\label{demonstration}
\end{figure}
Although the problem has attracted much attention of the research community, there are serious challenges, primarily from the intrinsic complexity of aging in the physical world and a shortage of labeled aging photograph data. Generally speaking, aging accuracy and identify permanence are commonly regarded as two crucial metrics to evaluate the quality of facial aging and rejuvenation in the recent literature~\cite{shu2015personalized,suo2010compositional,yang2016face}. 
\subsection{Related Work}

Early approaches were mainly based on the skin's anatomical structure and simulated the profile growth and facial muscle changes with respect to the elapsed time~\cite{ramanathan2008modeling}. Although these methods provided novel insights for facial aging synthesis, they are difficult to generalize for other tasks because of their complex modeling techniques. 

The later data-driven approaches can be roughly categorized into prototype-based methods~\cite{tiddeman2001prototyping,kemelmacher2014illumination,shu2015personalized} and physical model based ones~\cite{suo2010compositional,park2010age,suo2012concatenational}. By dividing training data into several disjoint age groups, prototype-based approaches learn a transformation over these age groups~\cite{burt1995perception,kemelmacher2014illumination}. Because they only consider the general aging mechanism, they are simple and fast. However, they neglect personalized information, causing them to generate unrealistic images. Although~\cite{shu2015personalized} utilized dictionary learning to estimate the age pattern of each age group from the corresponding sub-dictionary, this approach presents serious ghosting artifacts. Physical model based approaches, on the other hand, employ parametric models to simulate the aging mechanisms of the muscles~\cite{suo2012concatenational}, the wrinkle~\cite{ramanathan2008modeling,suo2010compositional}, the skin, and the skull of a particular individual~\cite{lanitis2002toward,ramanathan2006modeling}.  
Nevertheless, they suffer from a complex modeling procedure with high computational cost. Moreover, it is difficult for these approaches to collect a large ground truth face dataset, with a long time span of each individual, to model the subtle aging mechanism. 

Recently, the generative adversarial networks (GANs) have shown an impressive ability in generating synthetic images~\cite{goodfellow2014generative,gauthier2014conditional,radford2015unsupervised}, and facial aging and rejuvenation~\cite{wang2016recurrent,duong2017temporal,zhang2017age,yang2017learning}. For example,~\cite{wang2016recurrent} transformed faces across different ages smoothly by modeling the intermediate transition states in a RNN model. And~\cite{zhang2017age} proposed conditional adversarial autoencoder to simulate facial muscle sagging caused by aging. These approaches render faces with more appealing aging effects and less ghosting artifacts compared to the earlier methods. However, aging accuracy and identity permanence can hardly be achieved simultaneously. The reason is that they focus more on modeling facial transformation between age groups, where the age factor plays a dominant role while the identity information plays a subordinate role. Furthermore, learning facial aging between age groups does not allow the generation of facial images for an arbitrary age.

\subsection{Our Approach}

In this paper, we propose a novel GANs-based approach, named Conditional Multi-Adversarial Autoencoder with Ordinal Regression (CMAAE-OR), which combines the advantage of GANs in synthesizing visually plausible images and that of ordinal regression in accurate age estimation. Compared with existing methods, our method can simultaneously handle the identity permanence and aging accuracy better on facial aging and rejuvenation. Concretely, CMAAE-OR utilizes a convolutional encoder to extract a latent feature from an input face photograph, followed by projecting the feature onto the face manifold conditional on age through a deconvolutional generator. The encoder and the generator are trained with four parts, (1) an age-distance-based weighted squared Euclidean loss in the image space, (2) the identity loss to minimize the input-output distance by a latent feature representation, which embeds personalized characteristics from a pre-trained encoder, (3) the GAN loss that encourages generated faces to be indistinguishable from actual faces, and (4) the ordinal regression loss to force generated faces to exhibit desirable aging effect. These four parts simultaneously ensure that the resultant faces present desired aging effects and the identity properties remain stable. In contrast to the previous approaches that regarded an age group as a {\it conditional input}, the proposed method allows a {\it specific age input} and utilizes an age estimation technique to ensure the aging accuracy. Consequently, CMAAE-OR produces more photorealistic and aging accurate images, as shown in Figure~\ref{demonstration}.
The main {\bf contributions} are summarized as follows:
\begin{enumerate}
\item{The proposed method incorporates face verification and age estimation techniques to preserve identity permanence and achieve high aging accuracy. In addition, our framework accepts an arbitrary age as the conditional input, instead of a pre-defined discrete age group.}
\item{An age-distance-based weighted squared Euclidean loss in facial image space is utilized in our framework to emphasize the aging effect.}
\item{Experimental results illustrate the appealing performances of the proposed method in facial aging and rejuvenation. Besides, our method is robust against variations in pose, eyeglasses, and occlusion.}
\end{enumerate}


\section{The Method}\label{approach}

\begin{figure*}[ht!]
\centering
\includegraphics[width=0.98\linewidth,height=0.5\linewidth, clip=true, trim=0 180 0 0]{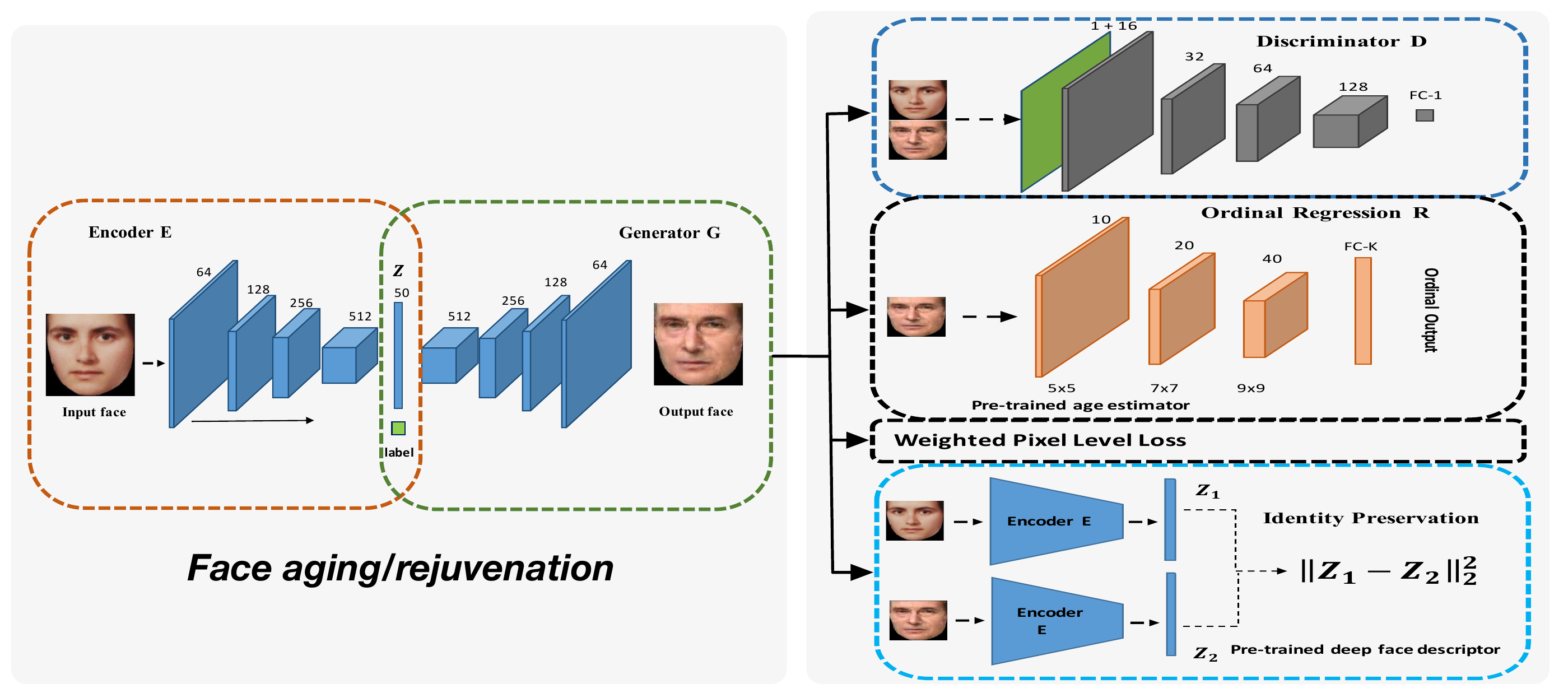}
\caption{Structure of the proposed CMAAE-OR framework for facial aging and rejuvenation. A convolutional encoder $E$ and a deconvolutional generator $G$ learn the age transformation in tandem. The training of CMAAE-OR incorporates four different losses: (1) the GAN loss that encourages generated faces to be indistinguishable from the provided actual faces, (2) the ordinal regression loss that makes generated faces exhibit desirable aging effect, (3) the weighted squared Euclidean loss in the image space that eliminates the input-output gap, and (4) the identity loss to minimizes the input-output distance by a latent features $z$, which embeds personalized characteristics.}
\label{structure}
\end{figure*}

We introduce the proposed Conditional Multi-Adversarial Autoencoder with Ordinal Regression (CMAAE-OR) in detail. Then the objective function of CMAAE-OR is described.

\subsection{Conditional Multi-Adversarial Autoencoder with Ordinal Regression}\label{CMAAE}
In our framework, the provided actual face image is first mapped to a latent vector through a convolutional encoder $E$. Then the vector is projected onto the face manifold conditional with a desired age through a deconvolutional generator $G$. The latent vector preserves personalized face features and the age controls facial aging or rejuvenation. To predict aging trend well and keep person-specific information stable, a compound training procedure with four different loss functions is employed. Specifically, (1) an age-distance-based weighted squared Euclidean loss in the image space is used for eliminating input-output gap, (2) the identity loss is for minimizing the input-output distance in a high-level feature representation which embeds the personalized characteristics, (3) the GAN loss is for encouraging generated faces to be indistinguishable from the actual faces, and (4) the ordinal regression loss is for ensuring the aging accuracy of the generated faces. The detailed structure of the CMAAE-OR is shown as Figure~\ref{structure}.

{\bf\noindent Encoder \& Generator}:
Facial aging and rejuvenation only require a forward pass through encoder $E$ and a generator $G$. The encoder $E$ maps the input face $x$ to a feature vector, {\it i.e.}, $E(x) = z\in\mathcal{R}^n$, where $n$ is the dimension of the face feature. Given $z$ and conditional age label $\ell$, the generator $G$ generates the output face $\hat x = G(z, \ell)$. Unlike existing GAN-related works~\cite{zhang2017age,yang2017learning}, the age label $\ell$ is a specific age with one dimension rather than an age group with a one-hot age label, so that a specific aged face can be generated.

{\bf\noindent Discriminator}:
According to the principle of conditional generative adversarial networks (cGANs)~\cite{mirza2014conditional}, the discriminator $D$ on face images forces the generator $G$ to yield more realistic faces. The goal of the generator $G$ is to confuse the discriminator $D$ through capturing the distribution of true face, whereas the optimization procedure of $D$ is to distinguish the natural face images from the ones generated by $G$. The risk function of optimizing this mini-max two-player game can be written as:
\begin{eqnarray}
\mathcal{V}(D, G) &=& \min_G\max_D\mathbb{E}_{x,\ell\sim p_{data}(x, \ell)}\log[D(x, \ell)] +\\ 
                  && \mathbb{E}_{x,\ell\sim p_{data(x,\ell)}}\log[1 - D(G(x, \ell))]\;,\nonumber
\label{eq_gan}
\end{eqnarray}
where $x$ denotes an actual face image following a certain distribution $p_{data}$ and $\ell$ is a conditional age label with one dimension. After the process converges, the distribution of the synthesized images $p_g$ is equivalent to $p_{data}$. Accordingly, the training process alternately minimizes the following equations:
\begin{eqnarray}
\mathcal{L}_{gan-d} &=& \mathbb{E}_{x,\ell\sim p_{data}(x, \ell)}[-\log D(x, \ell)] + \\
&&E_{x, \ell\sim p_{data}(x, \ell)}[-\log(1 - D(G(E(x), \ell)))]\;,\nonumber \\
\mathcal{L}_{gan-g} &=& \mathbb{E}_{x, \ell\sim p_{data}(x, \ell)}[\log(1 - D(G(E(x), \ell)))]\;.
\end{eqnarray}

{\bf\noindent Ordinal Regression}:
Once the aforementioned procedure is completed, an age estimation technique is used to help the generated image become more accurate in the age-level. In our case, a CNN-based ordinal regression method $R$ is introduced to estimate the age of the generated face image, because it has achieved a remarkable accuracy in the age estimation area~\cite{niu2016ordinal}. The regression loss for the generated image is written as 
\begin{equation}
\mathcal{L}_{regression} = \mathbb{E}_{x\sim p_{data}}{\Arrowvert R(G(E(x), \ell)) - \ell\Arrowvert}_2^2\;.
\end{equation}
Here $||\cdot||_2^2$ is the $L_2$ distance between feature representations. For more implementation details of ordinal regression $R$, readers are referred to~\cite{niu2016ordinal}.

{\bf\noindent Identity Preservation}:
Another core issue of facial aging and rejuvenation is to keep the person-dependent or person-specific properties consistent. By measuring the input-output distance in a proper feature space that is sensitive to the identify change and relatively robust to other variations, we incorporate an associate constraint into our proposed model. Specifically, we utilize an encoder $E$ (pre-trained by training dataset) to extract a latent vector $z$, which preserves personal identity in high-level feature representation for each face image. Then the identity loss can be written as
\begin{equation}
\mathcal{L}_{identity} = ||E(G(x, \ell)) -  E(x)||_2^2\;.
\end{equation}
The architecture of the pre-trained encoder is the same as the encoder $E$.

\subsection{Objective Function}\label{objective}
Besides the three aforementioned loss functions, an age-distance-based weighted squared Euclidean loss in the image space is adopted for further eliminating the input-output gap, {\it e.g.}, the color aberration. Because the original squared Euclidean loss may eliminate the facial aging effect to some degree, we proposed an age-distance-based weighted squared Euclidean loss to avoid this issue. The proposed weighting strategy is based on the intuition that the higher the gap in age, the larger the gap in input-output faces. The formulation of this loss can be written as
\begin{equation}
\mathcal{L}_{pixel} = \frac{1}{{\Delta \ell}\times W\times H\times C}||G(x, \ell) - x||_2^2\;,
\end{equation}
where $W$, $H$, and $C$ correspond to the shape of image $x$ ({\it e.g.} width, height, and channel) and $\Delta \ell = |\ell_{in} - \ell_{out}|$. Here $\ell_{in}$ and $\ell_{out}$ are the age labels of the input face and the generated face, respectively.

Finally, the objective functions for the generator $G$ and the discriminator $D$ are written as:
\begin{eqnarray}
\mathcal{L}_{G} &=& \lambda_p\mathcal{L}_{pixel} + \lambda_i\mathcal{L}_{identity} +\label{gan_g}\\ 
&& \lambda_g\mathcal{L}_{gan-g} + \lambda_r\mathcal{L}_{regression}\;,\notag 
\\
\mathcal{L}_{D} &=& \mathcal{L}_{gan-d}\label{gan_d}\;,
\end{eqnarray}
where $\lambda_p$, $\lambda_i$, $\lambda_g$ and $\lambda_r$ are the coefficients of pixel loss, identity loss, GAN loss, and ordinal regression loss, respectively. These coefficients are trade-offs between face aging accuracy and identity performance.

In our framework, we first pre-train a convolutional encoder $E$ as a face identity descriptor and a CNN-based ordinal regression $R$ as an age estimator based on the training dataset. Then $G$ and $D$ are trained alternately by Eqs.~\eqref{gan_g} and~\eqref{gan_d} until optimality reaches. Finally, $G$ learns the desired age transformation pattern and $D$ becomes a reliable discriminator.

\section{Experimental Results}\label{exp}
We perform a comprehensive comparison between our proposed approach and several published methods.
\begin{figure*}[!ht]
\centering
\includegraphics[width=0.98\linewidth,height=0.7\linewidth, clip=true, trim=0 0 80 0]{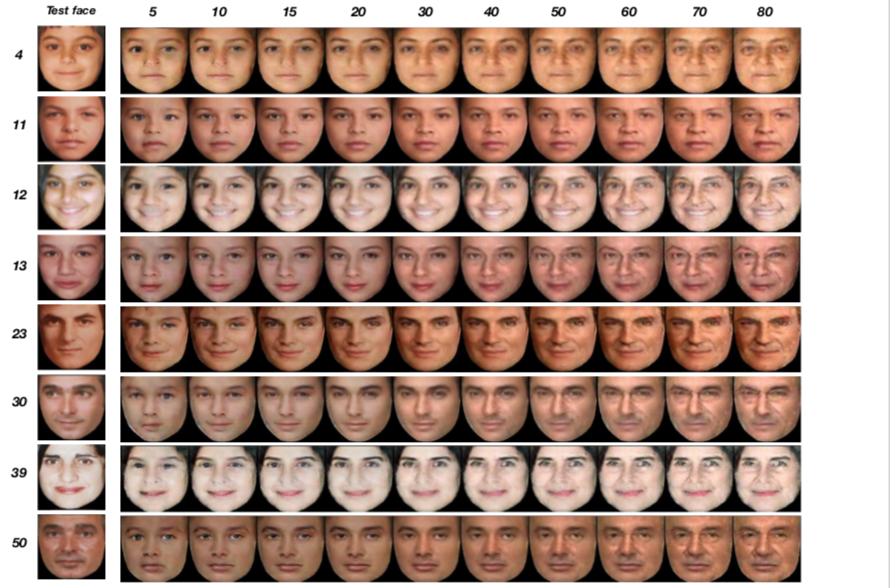}
\caption{The results of the proposed method on the FGNET dataset. The first column shows the provided faces. The other columns are our results for both facial aging and facial rejuvenation.}
\label{fgnet_result}
\end{figure*}
\subsection{Data Collection}
In our experiments, we utilized three datasets for training and evaluation: the MORPH dataset~\cite{ricanek2006morph}, the UTKFace dataset~\cite{zhang2017age}, and the FGNET dataset\footnote{\url{http://www.prima.inrialpes.fr/FGnet/}}.


The {\bf MORPH} is a large publicly available aging database, consisting of subject's ethnicity, height, weight, and gender. It contains 55,608 facial images and the age of each subject ranges from 16 to 77 years, with the average age being approximately 33 years. The {\bf UTKFace} is also a large-scale face dataset with long age span, ranging from 0 to 116 years. This dataset includes 23,709 facial images with annotations of age, gender, and ethnicity. The third facial aging dataset, the {\bf FGNET}, consisting of only 1,002 images of 82 subjects, is used for testing in our work. In addition, images from these three datasets cover a wide variations in eyeglasses, pose, facial expression, illumination, occlusion, etc.

In order to make the training phase effective, we align all the faces according to 68 landmarks in each face~\cite{kazemi2014one}. Each facial image is cropped to $128\times 128$ pixels.

\subsection{Implementation Details}
The architecture of the CMAAE-OR is constructed as in Figure~\ref{structure}. Specifically, we normalize four parameters into $[0, 1]$. They are (1) the pixel values of the input images, (2) the output $z$ of $E$ by using a sigmoid activation function, (3) the value of the label through dividing the maximal age of each dataset, and (4) the output of the network $G$ through using the sigmoid function. Furthermore, the desired age label is concatenated to $z$, forming the input of $G$. Based on our experiences, such a normalization helps the training process converge faster. In $E$, $G$, and $D$, the convolution of stride 2 is employed instead of pooling ({\it e.g.}, max pooling) because strided convolution is fully differentiable and allows the network to learn its own spatial downsampling~\cite{radford2015unsupervised}. Note that we do not use the batch normalization (BN) for $E$ and $G$ because it blurs personal features and makes the generated faces drift far away from inputs in testing. However, BN will make the framework more stable if it is applied on $D$. All intermediate layers of each block ({\it i.e.}, $E$, $G$, $D$, and $R$) use the ReLU as the activation function. Further, paddings are added to the layers to make the size of the input and the output identical.

The coefficients of four parameters $\lambda_p, \lambda_i, \lambda_g$, and $\lambda_r$ are set to 0.10, 1.00, 1.00, and 0.02 for the MORPH, and 0.50, 1.00, 1.00 and 0.01 for the UTKFace, respectively. At the training stage, we employ Adam~\cite{kingma2014adam} with the initial learning rate of $1\times 10^{-4}$ and the weight decay factor of $1\times 10^{-5}$. After the ordinal regression $R$ and the encoder $E$ are pre-trained, we alternatively update the discriminator $D$ with GAN loss and the generator $G$ with GAN loss, age estimation loss, pixel-level loss, and identity loss at every iteration. The networks are trained with a batch size of 100 for 200 epochs in total, which takes around 2.5 hours on four GTX 1080Ti GPUs.

\subsection{Performance Comparison}
\begin{figure}[!ht]
\centering
\includegraphics[width=1\linewidth,height=0.9\linewidth, clip=true, trim=0 20 110 0]{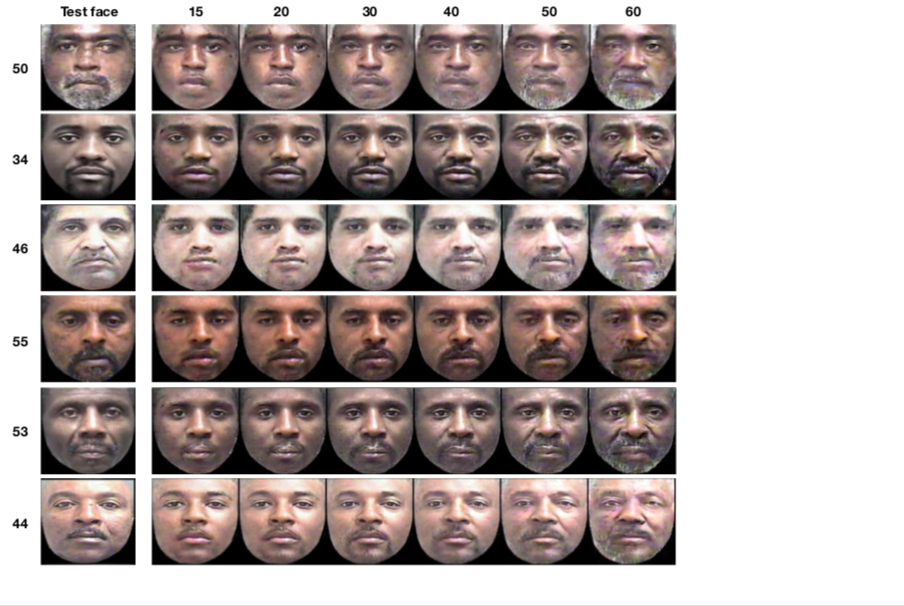}
\caption{The results of the proposed method on the MORPH dataset.}
\label{morph_resutl}
\end{figure}

{\bf\noindent Facial Aging and Rejuvenation}:
Given an input face and its target age label, CMAAE-OR generates the target age faces along the direction of facial aging or rejuvenation. We evaluate its performance for the FGNET and the MORPH datasets. For the MORPH, we randomly divide the whole dataset into the training set (80\%) and the testing set (20\%) without overlapping. For the FGNET, we utilize the UTKFace dataset as the training dataset and the FGNET as the testing set, following the setting in~\cite{zhang2017age}. The facial aging and rejuvenation results of our method on these two datasets are shown in Figures~\ref{fgnet_result} and~\ref{morph_resutl}, respectively. It can be seen that CMAAE-OR preserves the personal identity well even with a long age span and produces richer texture such as wrinkle in older faces, making older faces more realistic. 

\begin{figure}[!ht]
\centering
\includegraphics[width=0.98\linewidth,height=0.95\linewidth, clip=true, trim=0 20 150 0]{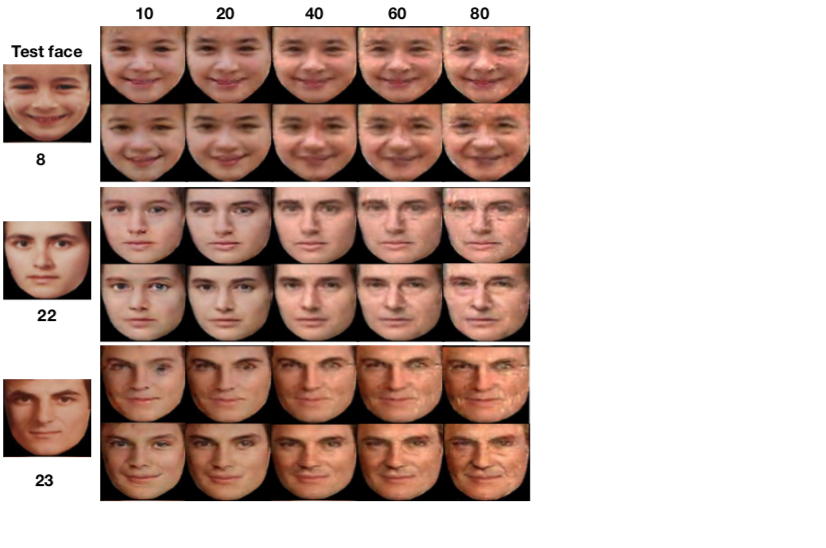}
\caption{Effect of the ordinal regression, which forces the age of the generated face to be closer to the given value. The first column shows the original faces, and their actual ages are marked below them. The other columns are generated faces through the proposed framework, without (the upper row) or with (the lower row) ordinal regression. The generated faces fall in five ages as indicated above the columns.}
\label{effect_r}
\end{figure}

{\bf\noindent Aging Accuracy and Effect of Ordinal Regression}:
To verify the accuracy of facial aging and rejuvenation of our method, an ordinal regression $R$ is trained as age estimator (the architecture of $R$ is shown in Figure~\ref{structure}), which achieves the remarkable performance for the MORPH and the FGNET datasets (see the left column of Table~\ref{MAE}). Then we compared the performances of generating facial images by the CMAAE-OR with and without the ordinal regression network $R$, to justify the effectiveness of the ordinal regression in our networks. The age estimation errors of synthesized images under these conditions are shown in the right column of Table~\ref{MAE}. It can be seen that the ordinal regression $R$ improves the accuracy of synthesized facial images remarkably. Further, Figure~\ref{effect_r} shows the effects of $R$ to synthesized images. It is clear that $R$ helps the framework generate aging face with high accuracy. Although the outputs without $R$ can present aging, its effect is subtle since $R$ brings age-level details to the generated face. The detailed results are illustrated in Table~\ref{MAE}. Note that other published work should have similar MAE as our method without $R$. The reason is that those methods evaluate MAE in discrete age groups. That can cause the 50-aged face and the 60-aged being placed in the same group, producing a large MAE value.  

\begin{table}[ht]
    \centering
    \caption{The comparisions between CMAAE-OR with oridnal regression ($R$) and CMAAE-OR without $R$. The performance is measured by the Mean Absolute Error (MAE) metric.}\label{MAE}
    \resizebox{\columnwidth}{!}{
    \begin{tabular}{|c|c|cc|}
    \hline
        \multirow{2}{*}{Dataset} & \multicolumn{1}{|c|}{Actual Faces} & \multicolumn{2}{|c|}{Synthesized Faces} \\ \cline{2-4}
        		 & Estimator $R$  & with $R$  				  & without $R$ \\ \hline
        MORPH    & $3.27\pm 0.02$ & {$\mathbf{1.48\pm 0.39}$} & $10.42\pm 0.73$ \\ \hline
        FGNET    & $4.58\pm 0.08$ & {$\mathbf{3.62\pm 0.54}$} & $19.91\pm 0.82$ \\ \hline
    \end{tabular}
    }
\vspace{-1em}
\end{table}

\begin{figure}[!ht]
\centering
\includegraphics[width=0.98\linewidth,height=0.6\linewidth, clip=true, trim=10 80 20 0]{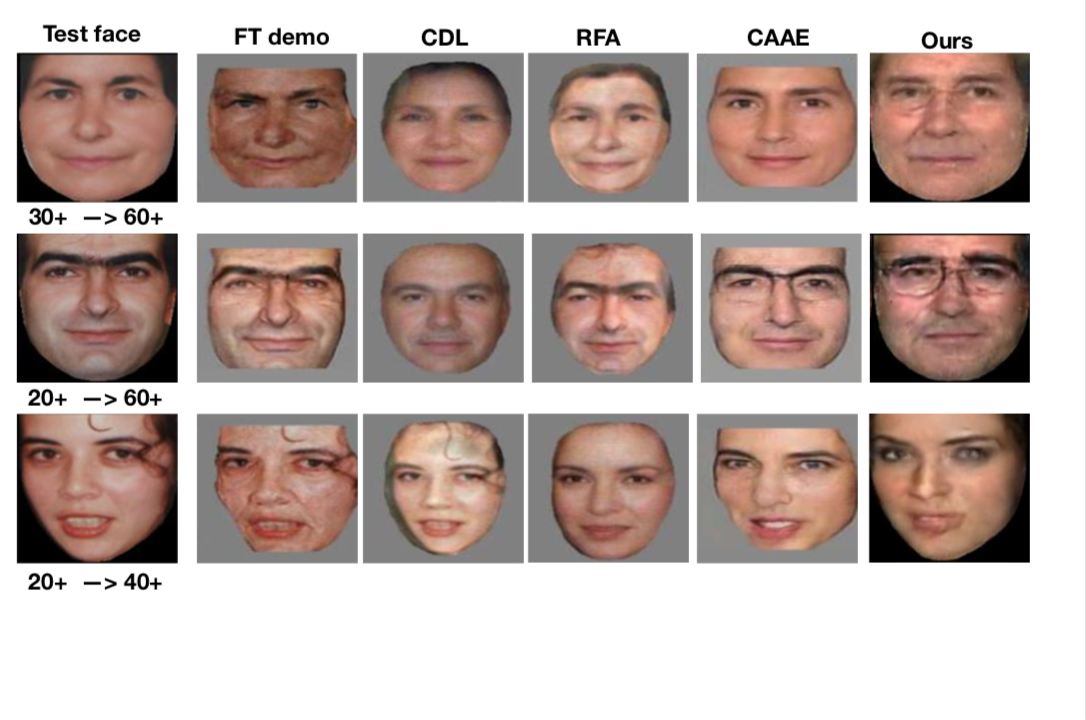}
\caption{Comparison to prior works FT demo, CDL, RFA, and CAAE.}
\label{compare_prior}
\end{figure}

{\bf\noindent Comparison with Prior Works}:
We compare our synthetic results with several representative prior works, including FT demo\footnote{Face Transformer (FT) demo.~\url{http://cherry.dcs.aber.ac.uk/transformer/}}, CDL: coupled dictionary learning~\cite{shu2015personalized}, RFA: recurrent facial aging~\cite{wang2016recurrent}, and CAAE: conditional adversarial autoencoder~\cite{zhang2017age}. As a prototype-based method, FT demo regards different age groups as aging prototypes to learn the aging pattern. Different from FT,  CDL utilized dictionary learning to estimate age pattern between age groups. And RFA transformed faces across different ages by modeling the intermediate transition states in a RNN model, while CAAE utilized a conditional adversarial autoencoder network to achieve a bidirectional face aging. For fair comparison, we choose the same faces with their works as our input, and directly cite their synthetic results, as most of prior works did. From the results of Figure~\ref{compare_prior}, it can be seen that the age changing of the synthetic face is not obvious in these prior works. In contrast, our method achieves facial aging with more clear changes. Moreover, our network simultaneously achieves age progression and regression in the same framework, and can generate facial image with an arbitrary age instead of a pre-defined discrete age group.

\begin{figure}[!ht]
\centering
\includegraphics[width=0.98\linewidth,height=0.45\linewidth, clip=true, trim=0 100 70 0]{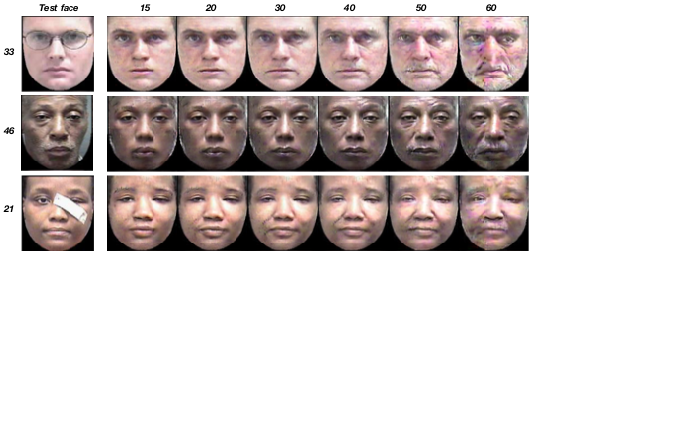}
\caption{Robustness to glasses, pose, and occlusion variations. The leftmost column shows the input faces, and the right columns are generated faces by CMAAE-OR from younger to older ages. In the first column, the first input face presents robustness to glasses, the second input shows only the face profile, and the last one is partially occluded by facial marks.
}
\label{robustness}
\end{figure}

{\bf\noindent Robustness}:
As aforementioned, the input images may have large variation in eyeglasses, pose, and occlusion. To demonstrate the robustness of the CMAAE-OR, we select the faces with eyeglass variation, non-frontal pose, and occlusion, respectively, as shown in Figure~\ref{robustness}. Note that the previous works~\cite{wang2016recurrent,kemelmacher2014illumination} often apply face normalization to alleviate the variation of pose and expression but they may still suffer from the occlusion issue. In contrast, the CMAAE-OR generates faces without the need of removing these variations, paving the way to robust performance in real-world applications.

\section{Conclusions}\label{con}
In this paper, a novel GANs-based method, named the Conditional Multi-Adversarial Autoencoder with Ordinal Regression (CMAAE-OR), is proposed to predict facial aging and rejuvenation.  It involves age estimation, face verification, and synthesis of visually  plausible images as well as eliminating the input-output gap by using a weighted pixel-level penalty. Different from the previous approaches, this method can simultaneously address face aging accuracy and identity permanence well. To our knowledge, what's more, it is the first time to generate an aged face with a specific age instead of a discrete age group. This can help face aging prediction obtain higher accuracy and finer estimate. Finally, experimental results on both face aging and rejuvenation demonstrate the effectiveness and robustness of the proposed method.


\bibliographystyle{named}
\newpage
\bibliography{ijcai18}

\end{document}